\def\year{2022}\relax
\documentclass[letterpaper]{article} 
\usepackage{aaai22}  
\usepackage{times}  
\usepackage{helvet}  
\usepackage{courier}  
\usepackage[hyphens]{url}  
\usepackage{graphicx} 
\usepackage{natbib}  
\usepackage{caption} 
\DeclareCaptionStyle{ruled}{labelfont=normalfont,labelsep=colon,strut=off} 
\usepackage{algorithm}
\usepackage{algorithmic,amssymb,amsmath}

\usepackage{subfigure}
\usepackage{multirow,makecell}
\usepackage{libertine}
\usepackage{newfloat}
\usepackage{listings}
\usepackage{color}
\floatstyle{ruled}
\newfloat{listing}{tb}{lst}{}
\floatname{listing}{Listing}
\title{Generating Natural Language Adversarial Examples \\through An Improved Beam Search Algorithm}
\title{Generating Natural Language Adversarial Examples \\through An Improved Beam Search Algorithm}
\author {
    Tengfei Zhao,\textsuperscript{\rm 1,2}
    Zhaocheng Ge, \textsuperscript{\rm 1,2}
    Hanping Hu, \textsuperscript{\rm 1,2}
    Dingmeng Shi, \textsuperscript{\rm 1,2}
}
\affiliations {
    \textsuperscript{\rm 1} School of Artificial Intelligence and Automation, Huazhong University of Science and Technology,\\
    \textsuperscript{\rm 2} Key Laboratory of Image Information Processing and Intelligent Control, Ministry of Education\\
    tenfee@hust.edu.cn, gezhaocheng@hust.edu.cn, hphu@mail.hust.edu.cn,m201972856@hust.edu.cn
}

\usepackage{bibentry}

\begin{document}

\maketitle

\begin{abstract}
The research of adversarial attacks in the text domain attracts many interests in the last few years, and many methods with a high attack success rate have been proposed. However, these attack methods are inefficient as they require lots of queries for the victim model when crafting text adversarial examples. In this paper, a novel attack model is proposed, its attack success rate surpasses the benchmark attack methods, but more importantly, its attack efficiency is much higher than the benchmark attack methods. The novel method is empirically evaluated by attacking WordCNN, LSTM, BiLSTM, and BERT on four benchmark datasets. For instance, it achieves a 100\% attack success rate higher than the state-of-the-art method when attacking BERT and BiLSTM on IMDB, but the number of queries for the victim models only is 1/4 and 1/6.5 of the state-of-the-art method, respectively.  Also, further experiments show the novel method has a good transferability on the generated adversarial examples.
\end{abstract}
\section{Introduction}
Adversarial samples\citep{43405,Szegedy2014}, by adding subtle and malicious perturbations to the original input, can easily deceive the deep neural network classifier and reduce its classification accuracy greatly. It reveals that deep neural networks are brittle when facing adversarial samples, but on the other hand, adversarial training\citep{akhtar2018threat,DBLP:conf/iclr/KurakinGB17}, which uses these generated adversarial examples for adversarial training, can improve the robustness of these deep neural networks. Adversarial examples are originated from the image field, and then various adversarial attack methods such as C\&W\citep{DBLP:conf/sp/Carlini017}, DEEPFOOL\citep{DBLP:conf/cvpr/Moosavi-Dezfooli16} have been produced. With the widespread use of DNN and CNN models in the NLP field, adversarial attacks in the natural language field have attracted more and more attention\citep{2019arXiv190207285W,DBLP:journals/access/XueYWZL20}. Compared with the image field, adversarial attacks on the text field are more challenging and harder to attack. This is mainly because (1) unlike images, the text is discrete (2) disturbances in the text are easily perceived, even if it is a character-level modification\citep{DBLP:journals/kbs/AlshemaliK20}. Adversarial attacks in the text field are divided into black-box attacks and white-box attacks. There are many white-box attacks and most of the white-box attacks are more effective\citep{2019arXiv190207285W,DBLP:journals/tist/ZhangSAL20}, However, in practice, the white-box attack strategy is not very suitable because white-box attacks require knowledge of the attacked model, which is often difficult to satisfy\citep{DBLP:conf/ccs/PapernotMGJCS17}. Black-box attacks are divided into three categories, character-level attacks, word-level attacks, and sentence-level attacks from the point of perturbation level. Wherein the word-level attacks are better, which can generate higher-quality adversarial texts\citep{DBLP:journals/corr/abs-1909-06723}. But most works
do not consider the runtime of the search algorithms. This has created a large, previously unspoken disparity in the runtimes of proposed works\citep{Yoo2020}. Furthermore, the benchmark word-level attack methods in a black-box scenario have the following disadvantages (1) their attack success rate can be further improved, especially in the field of short texts (2) they are quite inefficient, which requires much time to run the algorithms as they require lots of queries for the victim model when crafting text adversarial examples. Existing defense methods, such as PRADA\citep{DBLP:conf/eurosp/JuutiSMA19}, protect the DNN model by analyzing continuous API quires to resist adversarial attacks. Therefore, attack methods that need lots of queries are easier to defend.

The focus of our research in this paper is how to reduce queries for the attacked model while ensuring a higher attack success rate when crafting adversarial examples in a black-box scenario. Our main contributions are as follows:

\begin{enumerate}
	\item We propose a novel attack model in a black-box scenario, which can generate high-quality adversarial examples for tasks such as Sentiment Analysis and Natural Language Inference.
	\item We design novel attack strategies for long texts and short texts. Therefore, our attack model has a higher attack success rate and higher efficiency, which the model generates adversarial examples more efficiently than the previous models.
	\item The adversarial examples generated by our model with high semantic similarity, low perturbation, and good transferability.
	
\end{enumerate}

We organize a lot of experiments, and the experimental results show that our method can significantly shorten the running time and reduce the number of model queries when generating adversarial examples. In most of the experiments, the attack speed of our attack method is about 3-15 times faster than baseline methods. Besides, our method can ensure the attack success rate and the quality of adversarial examples. For instance, our attack method can achieve a 100\% attack success rate when attacking BERT/BiLSTM on the IMDB dataset, but the running time and the number of queries is 1/15-1/3 of the baseline methods. Our algorithm can achieve a 91.6\% attack success rate when attacking BERT on the SST-2 dataset, which is higher than all the baseline methods.

\section{Related Work}
 White-box attacks are all set in a white-box scenario, and they require accessing the attacked model and know the knowledge of the attack model. Since this restriction is difficult to satisfy, most investigations of the robustness of NLP models have been conducted using black-box adversarial methods\citep{DBLP:journals/kbs/AlshemaliK20}.

\citet{DBLP:conf/sp/GaoLSQ18} design a unique scoring mechanism based on classification confidence. Determine important tokens through this scoring mechanism and then perform character-level modification operations(i.e., swap, insert and delete). The black-box attack method of \citet{DBLP:conf/ndss/LiJDLW19} is similar to the process. \citet{DBLP:conf/emnlp/JiaL17} design a sentence-level attack method in a black-box scenario for reading comprehension systems, which by adding a sentence. \citet{DBLP:conf/acl/RenDHC19} comprehensively determine the replacement order of the replaced words, which are nouns, verbs, adjectives, adverbs, and entity words through the classification confidence and word saliency.  \citet{DBLP:conf/emnlp/AlzantotSEHSC18} design an attack method in a black-box scenario based on the Genetic algorithm, which randomly generates a set of adversarial texts for the replaced words, and iteratively select the adversarial texts with high confidence in the target classification, and perform crossover and mutation operations until the original classification change. \citet{DBLP:conf/acl/ZangQYLZLS20} design an attack method in a black-box scenario based on the particle swarm optimization algorithm. The basic idea of this method is similar to the \citet{DBLP:conf/emnlp/AlzantotSEHSC18}, but the attack success rate of this method is higher, and the generated adversarial examples quality is better.

Word-level attacks in a black-box scenario can be regarded as a combinatorial optimization problem. Word-level attacks based on word synonym substitution, represented by PWWS\citep{DBLP:conf/acl/RenDHC19} and SememePSO\citep{DBLP:conf/acl/ZangQYLZLS20}. The series of algorithms(greed-based) represented by PWWS require fewer model queries and fewer computing resources, but the attack success rate of such algorithms is not as good as those methods like SememePSO\citep{DBLP:journals/corr/abs-2009-06368}. The series of algorithms(population-based) represented by SememePSO are significantly more expensive than other algorithms. Because such population-based search methods like PSO, by constant iteration and optimization to solve the text discrete task and search the optimal solution, but because the convergence speed of the process is extremely slow and the search efficiency is inefficient, thus a large number of queries are required, they need more computing resources. The number of iterations and the size of the population are the main factors. 
In addition, the above models have a low attack success rate on short text datasets, and there is still room for improvement.

\section{Attack Design}
Before generating adversarial examples, we need to generate synonyms for those words which will be substituted; 
we use the Stanford-PoSTagger\footnote{\url{https://nlp.stanford.edu/software/tagger.shtml}} to tag all words in the sentence. Then we use the sememe-based Hownet\citep{dong2006hownet} and the synonym-based Wordnet\citep{DBLP:journals/cacm/Miller95}  to get the synonyms of content words. Content words are words that carry meanings and
consist mostly of nouns, verbs, adjectives, and adverbs\citep{DBLP:conf/acl/ZangQYLZLS20}.
We choose Hownet and Wordnet to generate synonyms because the substitution words generated by Hownet and Wordnet are richer and of higher quality. Finally, we use Sentiwordnet\citep{DBLP:conf/lrec/Esuli006} to calculate the semantic similarity between every content word's synonyms and itself, and sort its synonyms in descending order, which meaning the higher the similarity of a synonym, the higher the priority of being replaced.

To improve the efficiency and the success rate of the attack, we design the following innovations:
\begin{enumerate}
	\item We improve the base beam search algorithm based on the particularity of the adversarial text generation and apply it to the process of searching the solution.
	\item The size of the solution space for long texts and short texts is different, thus we propose different attack strategies.
	\item We design an effective scoring mechanism based on TF-IDF and neutral word replacement to determine the replacement order of content words.
	
\end{enumerate}

\subsection{Improved Beam Search Algorithm}
Beam search is a widely used approximate search algorithm. It strives to focus the search on the most promising path. The beam search can find the solution to the problem within the actual time and memory constraints, even if the search space is huge\citep{DBLP:conf/aips/ZhouH05}.
Beam search is widely used in various NLP tasks, such as speech recognition\citep{lowerre1976harpy}, neural machine translation\citep{DBLP:journals/corr/abs-1211-3711,DBLP:conf/ismir/Boulanger-LewandowskiBV13}, scheduling\citep{DBLP:conf/wsc/HabenichtM02}. When searching in discrete space and generating valid an adversarial example, although beam search cannot guarantee an optimal solution, it can efficiently get a valid adversarial example. The working principle of the beam search algorithm is that each iteration takes one content word and replaces it with its synonyms, and selects the K(beam width) optimal nodes for the next iteration. This method has obvious defects because if the initial selected K nodes are invalid, the negative impact will affect the subsequent process.

\begin{figure}[ht]
	\centering
	\includegraphics[scale=0.40]{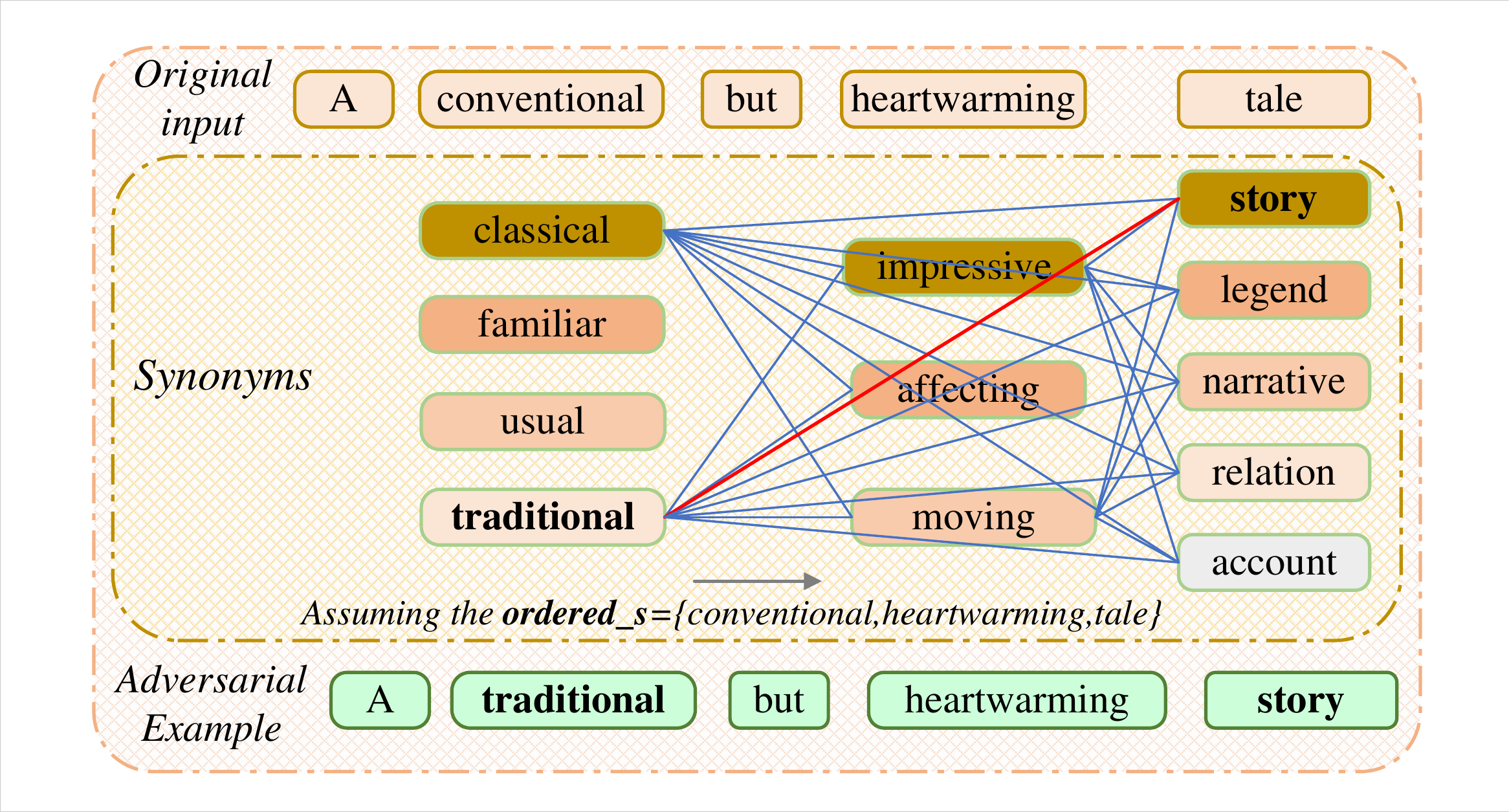}	
	\caption{An example showing the word substitute strategy and search strategy in word-level adversarial attacks on short sentences. Blue bold lines show all possible combinations, red bold line is the combination that makes the attack successful. The darker the color means that the synonym and the replaced content word are more similar in semantics, and the replacement priority of the synonym is also higher. }
	\label{fig:1}
\end{figure}

To solve this problem, we make the following improvement to the base beam search algorithm. Each iteration is no longer just the K nodes selected in the last iteration, but the K nodes selected in the last iteration and the nodes participate in the last iteration. The improved beam search algorithm not only makes the combination of synonyms more sufficient but also weakens the Butterfly Effect. The experiments prove that the improved beam search algorithm not only has a higher attack success rate but also greatly shortens the running time.

\subsection{Adversarial Example Search Algorithm}
For an original input, we formalize it as \( \mathbf{x}\), \( \mathbf{x}=w_1w_2\cdots w_n\), where \(w_i\) is the i-th word in the original input, \(n\) is the length of the original input, which is the number of words in the original input text. We formalize the real label as \(y_\mathbf{{true}}\), and the target label as \(y_\mathbf{{target}}\), where \(y_\mathbf{{true}}, y_{\mathbf{{target}}} \in \mathcal{Y}\), \(\mathcal{Y}\) is a set containing all labels. In the following, we will introduce our Word Substitution Strategy and Search Strategy separately.

\subsubsection{Word Substitution Strategy}
We create different word substitution strategies for long sentences and short sentences. Attacking short sentences is more challenging because these sentences length is short and there are fewer content words, the combination of synonyms is insufficient, and the solution space is small. We create a scoring algorithm based on neutral word replacement\citep{DBLP:conf/acl/PruthiDL19} and TF-IDF\citep{jones1972statistical}. When attacking short texts, we first use this sorting algorithm to sort the content words to determine their priority. But for long sentences, there are many content words, the combination of synonyms is rich, and the solution space is large, we no longer calculate the priority of content words, but take a synonym of each content word in turn and combine them together, until all synonyms of each content word have been taken out.

\begin{figure}[ht]
	\centering
	\includegraphics[scale=0.40]{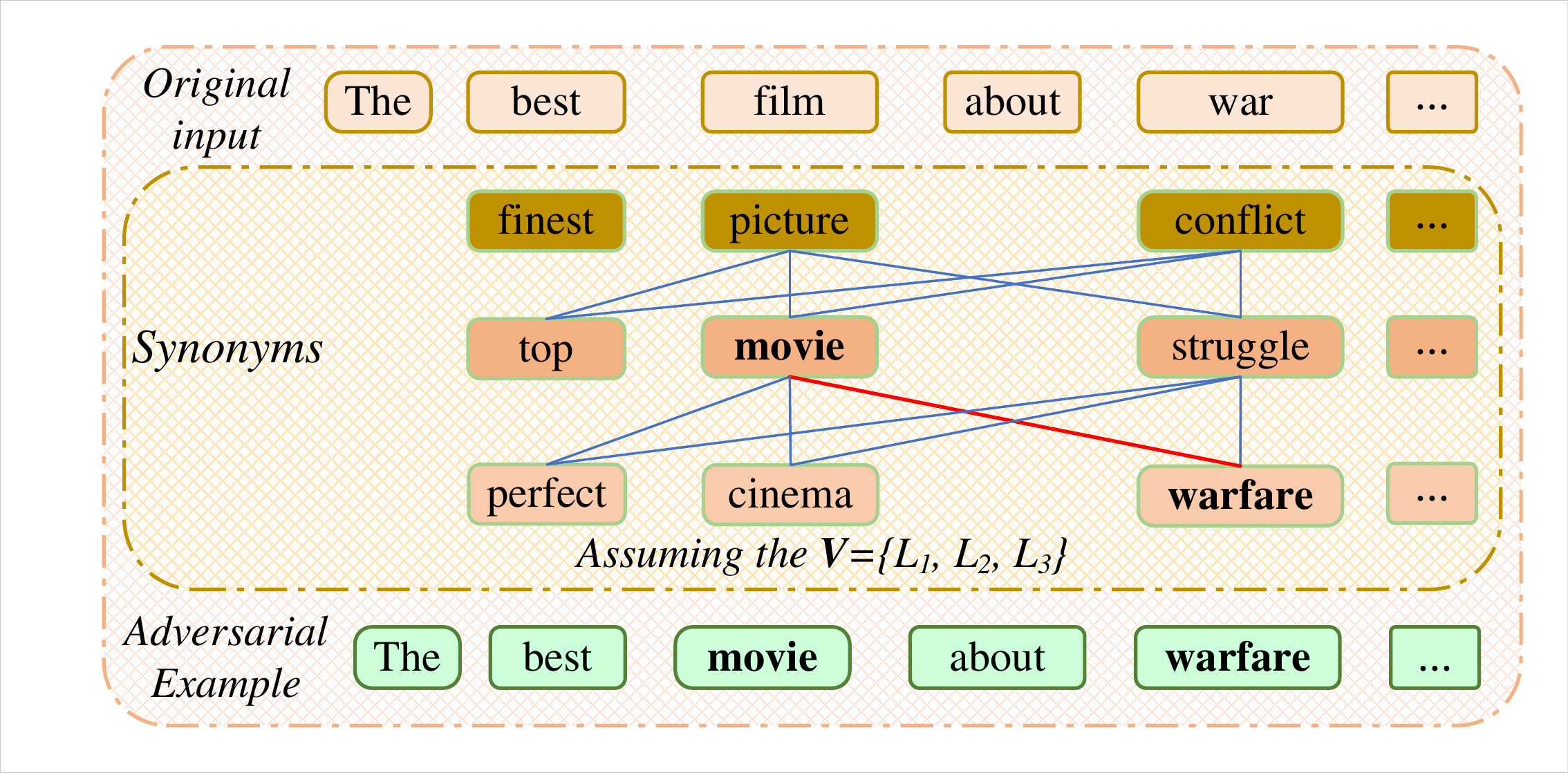}	
	\caption{An example showing the word substitute strategy and search strategy in word-level adversarial attacks on long sentences. Blue bold lines show all possible combinations, red bold line is the combination that makes the attack successful. The darker the color means that the synonym and the replaced content word are more similar in semantics, and the replacement priority of the synonym is also higher.}
	\label{fig:2}
\end{figure}

First, for an original input \(\mathbf{x}\) we abbreviate it as \(\mathbf{s},\mathbf{s}=s_1s_2\cdots s_m\), where \(s_i\) is the \(i\)-th content word of the original input text and \(m\) is the number of content words in the original input. We assume \(\mathbb{V}(s_{i})\) is the set of synonyms for \(s_i\), and synonyms in\(\mathbb{V}(s_{i})\) are sorted in descending order according to similarity by Sentiwordnet. If sentence \(\mathbf{x}\) is a short text then sort the content words \(\{s_1,s_2,\cdots s_m\}\) of \(\mathbf{x}\) according to the flowing formulae and get the sorted set \(\mathit{ordered\_s}\).

\begin{equation} \label{1}
	importance(s_i) =\psi(s_i)\times \phi({s_i}),
\end{equation}
\begin{equation} \label{2}
	\psi(s_i) =P(y_\mathbf{{true}}|\mathbf{s})-P(y_\mathbf{{true}}|\mathbf{s}_{i}^{\prime}),
\end{equation}
\begin{equation}\label{3}
	\phi({s_i})=\frac{e^{{tfidf(s_i)}}}{\sum_{k=1}^{m}e^{{tfidf(s_k)}}},
\end{equation}
\begin{equation}\label{4}
	tfidf(s_i)=tf_{s_i}\times \ln\big(\frac{N}{df_{s_i}+1}),
\end{equation}
where \(\mathbf{s}_{i}^{\prime}=s_1\cdots s_{i-1}\texttt{\small{UNK}}s_{i+1}\cdots s_m\), \texttt{\small{UNK}} represent those words that out of the vocabulary, \(tf_{s_i}\) is the number of occurrences of \(s_i\) in \(\mathbf{x}\),  \(df_{s_i}\) is the number of orignal inputs(training set) containing \(s_i\), \(N\) is total number of orignal inputs(training set). Inspired by \citet{DBLP:conf/acl/PruthiDL19} we use \texttt{\small{UNK}} to substitude \(s_i\), but for the SNLI dataset, there is no \texttt{\small{UNK}} due to its limited dictionary, we use "a" which has a similar distribution across classes.

However,if sentence \(\mathbf{x}\) is a long text, then we no longer sort the content words \(\{s_1,s_2,\cdots s_m\}\) of \(\mathbf{x}\), but combine the synonyms \(\{\mathbb{V}(s_{1}),\mathbb{V}(s_{2}),\cdots,\mathbb{V}(s_{m})\}\) of the content words according to the following formulae, and get \(\mathbb{V}\). 
\begin{equation} \label{5}
	\mathbb{V}=\big\{\mathit{L}_1,\mathit{L}_2,\cdots,\mathit{L}_{max}\big\}
\end{equation} 
\begin{equation} \label{6}
	\mathit{L}_i=\big\{\mathbb{V}(s_1)^i,\mathbb{V}(s_2)^i,\cdots,\mathbb{V}(s_m)^i\big\}
\end{equation} 
Where the value of \(max\) in Eq. (\ref{5}) is the maximum value in \(\{l_1,\cdots,l_m\}\) where \(l_d\) is the number of synonyms of \(d\)-th content word of \(\mathbf{x}\), \(\mathbb{V}(s_d)^i\) is the \(i\)-th(\(i<=l_i\)) synonym of the \(d\)-th content word \(s_d\).

\begin{table*}
	\centering
	\small
	\renewcommand{\arraystretch}{1.5}
	\begin{tabular}{c|c|c|c|c|c|c}
		\hline
		\textbf{Dataset} & \textbf{Task} & \textbf{\#Classes} & \textbf{\#Train} & \textbf{\#Dev} & \textbf{\#Test} & \textbf{\#Avg words}\\
		\hline\hline
		IMDB & Sentiment Analysis & 2 & 25,000 & 0 & 25,000 & 234 \\
		SST-2 & Sentiment Analysis & 2 &6920 & 872 & 1821 & 17\\
		YELP & Sentiment Analysis & 2 & 560,000 & 0 & 38,000 & 152\\
		SNLI & NLI & 3 & 550,152 & 10,000 & 10,000 & 8\\
		\hline
	\end{tabular}
	\caption{\label{tabel:1}
		Details of datasets. "\#Classes" denotes the number of classifications, "\#Train", "\#Dev", and "\#Test" respectively denotes the number of samples of the training set, validation set, and test set. "\#Avg words" denotes the average sentence length of the dataset.}	
\end{table*}

\subsubsection{Search Strategy}
In the following, we will introduce the search strategy of our algorithm, short sentences use the improved beam search algorithm to search, long sentences first use the base beam search algorithm to search. We assume \(S^*=\{\mathbf{x}\}\), \(i=1\), \(l_d\) is the number of synonyms of \(d\)-th content word of \(\mathbf{x}\), \(len(S^*)\) is the number of elements in \(S^*\), \(n\) is the sentence length of \(\mathbf{x}\), and \(len(\mathbb{L}_i)\) is the number of elements in \(\mathbb{L}_i\). 

If sentence \(\mathbf{x}\) is a short text then take the \(i\)-th content word from \(\mathit{ordered\_s}\), assuming that the taken content word is \(s_d\). For each element in \(S^*\), replace its \(d\)-th content word with the following Eq. (\ref{7}) and get \(S^\prime\). 
\begin{equation} \label{7}
	\mathbf{x}^\prime=w_1,w_2,\cdots,s_d^\prime,\cdots,w_n
\end{equation} 
where \(s_d^\prime\ \in \mathbb{V}(s_d)\), each element in \(S^*\) will get \(l_d \times\) \(\mathbf{x}^\prime\), then we merge all \(\mathbf{x}^\prime\) into \(S^\prime\), finally the number of elements in \(S^\prime\) is \(len(S^*) \times l_d\). 

However, if sentence \(\mathbf{x}\) is a long text then take the \(\mathbb{L}_i\) from \(\mathbb{V}\). For each element in \(S^*\), replace its content words with the following Eq. (\ref{8}) and get \(S^\prime\). 

\begin{equation} \label{8}
	\mathbf{x}^\prime=w_1,w_2,\cdots,v,\cdots,w_n
\end{equation} 
where \(v \in \mathbb{L}_i\), each element in \(S^*\) will get \(len(\mathbb{L}_i) \times\) \(\mathbf{x}^\prime\), then we merge all \(\mathbf{x}^\prime\) into \(S^\prime\), finally the number of elements in \(S^\prime\) is \(len(S^*)\times len(\mathbb{L}_i)\).

If the following Eq. (\ref{9}) is true, then the attack is successful. Otherwise,  update \(S^*\) and \(i\) according to Eq. (\ref{10}), repeat the above process. If all the elements in \(\mathit{ordered\_s}\) or \(\mathbb{V}\) are taken out but the attack still not successful, then the attack fails.
\begin{equation} \label{9}
	\begin{split} 
		\arg \max _{y_i \in \mathcal{Y}} P(y_i|\mathbf{x}^\prime)=y_{\mathbf{target}}
		\\
		\mathbf{x}^\prime=\arg \max_{\mathbf{x}^\prime \in S^\prime}P(y_{\mathbf{target}}|\mathbf{x}^\prime)
	\end{split} 
\end{equation}
where \(\mathcal{Y}\) is a set containing all labels.
\begin{equation}\label{10}
	\begin{split}
		S^* &=
		\begin{cases}
			select(S^\prime,K)+S^*,&short\text{ }sentence\\
select(S^\prime,K), &long\text{ }sentence\\
		\end{cases}\\
		i&=i+1	
	\end{split}
\end{equation}

where function \(select\) first sorts the elements in \(S^\prime\) from large to small according to the value of \(P(y_{\mathbf{target}}|\mathbf{x}^\prime)\), and then chooses the top K elements from sorted \(S^\prime\). 

Attacks on the short sentences use the improved beam search algorithm, suppose K is 2, as shown in Figure \ref{fig:1}. Attacks on the long sentences first use the base beam search algorithm, suppose K is 2, as shown in Figure \ref{fig:2}. But if the base beam search algorithm attack fails, then use the improved beam search algorithm to attack. The reason for this is that long sentences are easy to attack; only a fraction of samples need to be attacked by the improved beam search algorithm. The strategy of combining the improved beam search algorithm with the normal beam search algorithm can maximize the attack success rate while ensuring high attack efficiency.
\begin{table*}[ht]
	\centering
	\resizebox{\textwidth}{20mm}{		
		\renewcommand{\arraystretch}{1.5}
		\begin{tabular}{c||c|c|c|c|c|c|c|c|c|c|c|c|c|c|c|c}
			\hline
			\multirow{3}*{\makecell[c]{\textbf{Attack} \\ \textbf{Algorithm}}} & \multicolumn{4}{c|}{\textbf{BiLSTM}} & \multicolumn{4}{c|}{\textbf{LSTM}}& \multicolumn{4}{c|}{\textbf{WordCNN}} & \multicolumn{4}{c}{\textbf{BERT}} \\\cline{2-17}
			& IMDB & SST-2 & YELP & SNLI& IMDB & SST-2 & YELP & SNLI & IMDB & SST-2 & YELP & SNLI& IMDB & SST-2 & YELP & SNLI\\ 
			\cline{2-17}
			&88.89&83.30&95.98&83.77&88.92&83.99&95.50&81.36&88.23&82.56&94.50&84.26&91.58&90.55&96.71&89.39\\
			\hline\hline
			PWWS & 98.9&85.8&92.2&44.3&98.0&85.8&89.4&44.0&99.3&$\mathbf{86.6}$&96.8&39.6&94.7&79.9&92.5&45.1\\
			\hline
			Genetic & 99.4&86.5&97.4&65.1&98.8&79.2&97.9&58.8&99.4&80.2&96.7&59.7&99.8&87.1&94.8&64.7\\
			\hline
			SememePSO & 99.8&92.7&99.0&75.5&$\mathbf{99.8}$&85.3&99.0&69.3&99.6&85.8&97.9&69.7&99.8&90.9&$\mathbf{94.9}$&78.4\\
			\hline
			Ours & $\mathbf{100}$&$\mathbf{93.6}$&$\mathbf{99.2}$&$\mathbf{77.2}$&$\mathbf{99.8}$&$\mathbf{86.8}$&$\mathbf{99.3}$&$\mathbf{72.0}$&$\mathbf{99.7}$&86.1&$\mathbf{98.4}$&$\mathbf{71.6}$&$\mathbf{100}$&$\mathbf{91.6}$&94.4&$\mathbf{79.7}$\\
			\hline
		\end{tabular}
	}
	\caption{\label{tabel:2}
		The attack success rate(\%) of different attack methods. Row 3 is the classification accuracy rate(\%) of the victim model on four datasets.}	
\end{table*}

\section{Experiments}
In this section, we organize a series of comparative experiments to evaluate our attacking method on multiple datasets and victim models. Following \citep{DBLP:conf/acl/ZangQYLZLS20}, the tasks we set for the experiments are Sentiment Analysis and Natural Language Inference. All the experiments are conducted on a server with two Intel
5218R CPUs running at 2.10GHz, 128 GB memory, and a GeForce RTX 3090 GPU card.

\subsection{Datasets}

We choose four benchmark datasets, which are IMDB, SST-2, YELP, and SNLI.

IMDB\citep{DBLP:conf/acl/MaasDPHNP11}, SST-2\citep{DBLP:conf/emnlp/SocherPWCMNP13}, and YELP\citep{DBLP:conf/nips/ZhangZL15} are all binary classfication datasets for Sentiment Analysis, labeled as positive or negative.
SNLI\citep{DBLP:conf/emnlp/BowmanAPM15} is a dataset for Natural Language Inference, which can be categorized into three classes: entailment, contradiction, and neutral. Table \ref{tabel:1} lists the details of these datasets. 

\subsection{Victim Models}
We choose four widely-used models. Bi-directional LSTM(BiLSTM) with a max pooling layer and its hidden states are 128-dimensional. LSTM with a fully connected layer and its hidden states are 128-dimensional. Word-based CNN(WordCNN)\citep{DBLP:conf/emnlp/Kim14} with a max pooling layer, two fully connected layers, and a convolutional layer consisting of 250
filters of kernel size 3. The above three models are used for IMDB dataset, SST-2 dataset, and YELP dataset, but for SNLI dataset, the victim models are from \citet{DBLP:conf/emnlp/ConneauKSBB17}. Besides, we also choose the BERT-Base-Uncased(BERT)\citep{DBLP:conf/naacl/DevlinCLT19} model as the victim model, and this model is used on all four datasets. Row 3 in Table \ref{tabel:2} shows the classification accuracy of these models.

\subsection{Baseline Attack Algorithms}
We choose three recent open-source adversarial attack algorithms as our baseline methods. All baseline methods are word-level attacks in a black-box scenario. Synonyms used in all methods are built by Hownet and WordNet and de-duplicate by Sentiwordnet. 

We choose PWWS\citep{DBLP:conf/acl/RenDHC19} as our first baseline method. PWWS is a greddy-based algorithm that requires fewer model queries and fewer computing resources, and the attacking effectiveness is stronger than other greddy based algorithms. We choose SememePSO\citep{DBLP:conf/acl/ZangQYLZLS20} and Genetic\citep{DBLP:conf/emnlp/AlzantotSEHSC18} as our other two baseline methods. They are all optimization-based algorithms, optimization-based algorithms are more effective than greddy-based algorithms, and the SememePSO attacking effectiveness is stronger than the Genetic attack method. The attack speed of SememePSO is faster, the attack success rate is higher, and the quality of the generated adversarial samples is better. Thus, so far the SememePSO is the state-of-the-art attack method.

\subsection{Experimental Settings}

For our algorithm, the value of beam width(\(K\)) in Eq. (\ref{10}) is 3, this value is obtained by experiment in Section Further Analysis. For the baseline algorithms, we set their hyperparameters to the values suggested in their papers.

To improve evaluation efficiency, we following \citep{DBLP:conf/acl/ZangQYLZLS20}. For each dataset, we randomly take out 1000 samples that can be correctly classified from its test set as the original input to the attack algorithms, and if the modification rate of the generated adversarial example more than 25\%, it is deemed that the attack fails. But different from \citet{DBLP:conf/acl/ZangQYLZLS20}, we restrict the length of these 1000 samples to 10-200. 

\subsection{Evaluation Metrics}
Following \citep{DBLP:conf/acl/ZangQYLZLS20}, we comprehensively evaluate the performance of the attack algorithms from the following metrics.

(1)The attack success rate is the proportion of adversarial samples that can successfully change the classification of the classifier to the target label, and following \citep{DBLP:conf/acl/ZangQYLZLS20} the modification rate is less than 25\%. For attack algorithms, try to make the attack success rate as high as possible.

(2)The number of queries refers to the number of times the attack algorithm needs to access and query the victim model to craft an adversarial sample. The fewer the number of queries, the lower the dependence of the attack algorithm on the model and the shorter the running time.

(3)The modification rate is the average modification rate of successfully attacked adversarial examples. The lower the average modification rate, the smaller the disturbance to the original inputs.

(4)The grammaticality is the average grammatical error increase rate of successfully attacked adversarial examples. The lower the grammaticality, the higher the quality of the adversarial examples crafted by the attacking algorithm. We use Language Tool\footnote{\url{https://www.languagetool.org}} to calculate the grammaticality.

(5)The perplexity reflects the degree of confusion of a sentence; the lower the perplexity, the more natural the adversarial examples. We use the GPT-2\citep{radford2019language} to calculate the perplexity.

\begin{table*}[ht]
	\centering
	\resizebox{\textwidth}{50mm}{
		\renewcommand{\arraystretch}{1.7}
		\begin{tabular}{cc||c|c|c|c||c|c|c|c||c|c|c|c||c|c|c|c}
			\hline
			\multirow{2}*{\makecell[c]{\textbf{Victim} \\ \textbf{Model}}}&\multirow{2}*{\makecell[c]{\textbf{Attacking} \\ \textbf{Algorithm}}} & \multicolumn{4}{c||}{\textbf{IMDB}} & \multicolumn{4}{c||}{\textbf{SST-2}}& \multicolumn{4}{c||}{\textbf{YELP}} & \multicolumn{4}{c}{\textbf{SNLI}} \\\cline{3-18}
			& & \%M & \%I & \%P & Q & \%M & \%I & \%P & Q & \%M & \%I & \%P & Q& \%M & \%I & \%P & Q\\ 
			\cline{3-17}
			\hline\hline
			\multirow{4}*{BiLSTM}&PWWS & 3.52&0.14&$\mathbf{135}$&2975&8.80&0.49&640.2&302&6.05&0.17&210.3&2010&10.9&1.21&341.3&$\mathbf{201.9}$\\\cline{3-18}
			&Genetic & 4.47&0.21&149&10449&9.88&0.48&657.6&495&6.48&0.21&209.8&8335&10.6&0.97&344.7&455.9\\ \cline{3-18}
			&SememePSO & $\mathbf{3.32}$&$\mathbf{0.11}$&$\mathbf{135}$&4657&$\mathbf{8.29}$&$\mathbf{0.26}$&638.6&286&$\mathbf{5.15}$&$\mathbf{0.12}$&$\mathbf{182.4}$&3865&$\mathbf{9.83}$&1.13&$\mathbf{313.1}$&403.1\\ \cline{3-18}
			&Ours & 3.83&0.13&137&$\mathbf{718}$&9.25&0.33&$\mathbf{596.4}$&$\mathbf{254}$&5.71&0.13&195.8&$\mathbf{566}$&11.9&$\mathbf{0.74}$&342.9&398.4\\ 
			\hline\hline
			\multirow{4}*{LSTM}&PWWS & 3.55&0.17&135.3&2974&9.46&0.21&616.8&299&6.77&$\mathbf{0.14}$&228&2012&10.6&1.16&375.9&$\mathbf{199.6}$\\\cline{3-18}
			&Genetic & 4.30&0.24&143.1&10882&10.87&0.23&669.3&475&6.75&0.21&209&9308&10.7&1.11&422.6&476.7\\ \cline{3-18}
			&SememePSO & $\mathbf{3.20}$&$\mathbf{0.12}$&$\mathbf{131.1}$&4619&$\mathbf{9.20}$&0.11&640.8&330&$\mathbf{5.45}$&0.15&$\mathbf{196}$&4219&$\mathbf{10.0}$&1.26&397.0&494.3\\ \cline{3-18}
			&Ours & 3.63&0.16&132.9&$\mathbf{692}$&9.80&$\mathbf{0.05}$&$\mathbf{573.3}$&$\mathbf{286}$&5.91&0.17&201&$\mathbf{609}$&12.0&$\mathbf{0.80}$&$\mathbf{364.7}$&466.3\\ 
			\hline	\hline	
			\multirow{4}*{WordCNN}&PWWS & 3.65&0.19&$\mathbf{132.9}$&2977&9.11&0.28&628.3&298&$\mathbf{6.17}$&$\mathbf{0.10}$&223.8&2048&10.6&1.55&376.9&$\mathbf{201.6}$\\\cline{3-18}
			&Genetic &5.32&0.22&152.7&11992&10.21&0.25&640.4&424&8.03&0.17&254.3&9828&10.4&1.08&360.3&597.0\\ \cline{3-18}
			&SememePSO &$\mathbf{4.11}$&$\mathbf{0.17}$&139.0&5746&$\mathbf{8.67}$&$\mathbf{0.20}$&612.9&286&6.60&0.15&$\mathbf{221.8}$&4865&$\mathbf{9.62}$&1.33&$\mathbf{334.5}$&559.5\\ \cline{3-18}
			&Ours &4.48&$\mathbf{0.17}$&142.9&$\mathbf{872}$&9.57&$\mathbf{0.20}$&$\mathbf{582.3}$&$\mathbf{276}$&7.26&0.23&237.0&$\mathbf{866}$&12.4&$\mathbf{0.84}$&376.3&465.8\\ 
			\hline\hline
			\multirow{4}*{BERT}&PWWS & 3.97&0.08&147.7&3009&8.43&0.30&614.8&$\mathbf{307}$&5.64&0.06&215.7&2006&10.7&1.04&$\mathbf{339.4}$&$\mathbf{203.1}$\\\cline{3-18}
			&Genetic & 3.60&0.10&138.8&6914&8.18&0.30&629.3&356&6.06&$\mathbf{0.02}$&213.9&6917&10.0&0.97&368.3&379.6\\ \cline{3-18}
			&SememePSO & $\mathbf{3.38}$&$\mathbf{0.08}$&137.9&5246&$\mathbf{7.64}$&0.27&593.8&392&$\mathbf{5.22}$&0.07&$\mathbf{199.0}$&4295&$\mathbf{9.59}$&0.72&383.1&328.5\\ \cline{3-18}
			&Ours & 3.69&0.16&$\mathbf{136.4}$&$\mathbf{1313}$&8.58&$\mathbf{0.25}$&$\mathbf{581.3}$&403&6.04&0.22&209.4&$\mathbf{1634}$&12.1&$\mathbf{0.63}$&394.5&389.3\\ 
			\hline
		\end{tabular}
	}
	\caption{\label{tabel:3}
		Evaluation results of adversarial example quality crafted by different attack methods. "\%M" is the modification rate, "\%I" is the grammatical error increase rate, "\%P" is the language model perplexity, and "Q" is the number of queries.}	
\end{table*}

\subsection{Experimental Results}

\subsubsection*{Attack Success Rate}

The attack success rate of each attack model is listed in Table \ref{tabel:2}, where the third row of Table \ref{tabel:2} is the classification accuracy of the victim model on four datasets. Experimental results show that our attack method can achieve an excellent attack success rate on most victim models on four datasets. It achieves a 100\% attack success rate when attacking BiLSM/BERT on IMDB. 
Compare with the baseline methods, the attack success rate of our attack method is higher than the baseline methods, especially on short sentences SST-2/SNLI is significantly higher than the baseline attack methods.

\begin{table*}[ht]
	\centering
	\begin{tabular}{m{12.2cm}||c}
		\hline
		\multirow{2}*{\textbf{Examples}} & \multirow{2}*{\textbf{Prediction}}\\	
		& \\ 	
		
		
		\hline
		I thought the movie \textcolor{green}{\emph{was}}(\textcolor{red}{\emph{existed}}) great I thought Kristine foulmouthed was \textcolor{green}{\emph{great}}(\textcolor{red}{\emph{ideal}}) and was glad to see her move on into some more interesting roles I even overlook the fact that the print I have wasn't quite put \textcolor{green}{\emph{back}}(\textcolor{red}{\emph{backward}}) together correctly but who cares.& \textbf{Postive}{$\Rightarrow$}\textbf{Negative}\\
		
		\hline
		A minor picture with a major identity crisis it's sort of true and it's sort of \textcolor{green}{\emph{bogus}}(\textcolor{red}{\emph{false}}) and it's ho all the way through.&\textbf{Negative}{$\Rightarrow$}\textbf{Postive}\\ 
		\hline
	
		It might be tempting to regard Mr and his collaborators as oddballs but Mr Earnhart's the \textcolor{green}{\emph{charming}}(\textcolor{red}{\emph{tempting}}) movie \textcolor{green}{\emph{allows}}(\textcolor{red}{\emph{leaves}}) us to see them finally as artists.&\textbf{Postive}{$\Rightarrow$}\textbf{Negative}\\ 
		\hline
		\textbf{Premise: }A guy with long hair is making a sand sculpture.&\multirow{2}*{\textbf{Neutral}{$\Rightarrow$}\textbf{Entailment}}\\
		\textbf{Hypothesis: }A guy is in a sandcastle building \textcolor{green}{\emph{competition}}(\textcolor{red}{\emph{match}}).&\\
		\hline
		
		\textbf{Premise: }A man in a blue t-shirt is taking a picture while a woman with an umbrella walks behind him.&\multirow{2}*{\textbf{Contradiction}{$\Rightarrow$}\textbf{Entailment}}\\
		\textbf{Hypothesis: }There is a \textcolor{green}{\emph{dog}}(\textcolor{red}{\emph{canine}}) doing the moonwalk in the background.&\\
		\hline
	\end{tabular}
	\caption{\label{tabel:4}
		The adversarial samples generated by our attack model on BERT. The content words are green and substituted(synonym) words are red and in brackets.}
\end{table*}

\begin{table}[ht]
	\centering
	\resizebox{\columnwidth}{30mm}{
		\renewcommand{\arraystretch}{1.3}
		\begin{tabular}{cc||c|c|c|c}
			\hline
			\multirow{2}*{\makecell[c]{\textbf{Victim} \\ \textbf{Model}}}&\multirow{2}*{\makecell[c]{\textbf{Attacking} \\ \textbf{Algorithm}}} & \multirow{2}*{\textbf{IMDB}} & \multirow{2}*{\textbf{SST-2}}& \multirow{2}*{\textbf{YELP}} & \multirow{2}*{\textbf{SNLI}} \\	
			&&&&\\\cline{3-6}		
			\cline{3-5}
			\hline
			BiLSTM&\multirow{1.8}*{SememePSO} & \multirow{1.8}*{13.72}&\multirow{1.8}*{$\mathbf{31.28}$}&\multirow{1.8}*{11.81}&\multirow{1.8}*{$\mathbf{52.95}$} \\ 
			{$\Downarrow$}&\multirow{2}*{Ours} & \multirow{2.2}*{$\mathbf{15.40}$}&\multirow{2.2}*{31.08}&\multirow{2.2}*{$\mathbf{11.99}$}&\multirow{2.2}*{49.25} \\ 
			{DistilBERT}&&&& \\
			
			\hline
			LSTM&\multirow{1.8}*{SememePSO} & \multirow{1.8}*{$\mathbf{15.73}$}&\multirow{1.8}*{$\mathbf{37.63}$}&\multirow{1.8}*{11.21}&\multirow{1.8}*{49.63} \\ 
			{$\Downarrow$}&\multirow{2}*{Ours} & \multirow{2.2}*{15.13}&\multirow{2.2}*{35.71}&\multirow{2.2}*{$\mathbf{12.48}$}&\multirow{2.2}*{$\mathbf{50.97}$} \\ 
			{DistilBERT}&&&& \\ 
			\hline
			WordCNN&\multirow{1.8}*{SememePSO} & \multirow{1.8}*{21.88}&\multirow{1.8}*{$\mathbf{32.58}$}&\multirow{1.8}*{$\mathbf{13.58}$}&\multirow{1.8}*{$\mathbf{50.50}$} \\ 
			{$\Downarrow$}&\multirow{2}*{Ours} & \multirow{2.2}*{$\mathbf{22.96}$}&\multirow{2.2}*{32.28}&\multirow{2.2}*{13.00}&\multirow{2.2}*{49.54} \\ 
			{DistilBERT}&&&& \\ 
			\hline
			BERT&\multirow{1.8}*{SememePSO} & \multirow{1.8}*{$\mathbf{33.96}$}&\multirow{1.8}*{$\mathbf{55.77}$}&\multirow{1.8}*{26.34}&\multirow{1.8}*{$\mathbf{51.28}$} \\ 
			{$\Downarrow$}&\multirow{2}*{Ours} & \multirow{2.2}*{32.80}&\multirow{2.2}*{51.96}&\multirow{2.2}*{$\mathbf{30.50}$}&\multirow{2.2}*{50.07} \\ 
			{DistilBERT}&&&& \\
			\hline
			
		\end{tabular}
	}
	\caption{\label{tabel:5}
		Transferability of adversarial examples crafted by SememePSO and our method. Higher attack success rate(\%) reflects higher transferability. }	
\end{table}

\subsubsection*{Attack Efficiency}

\begin{figure}
	\centering
	\includegraphics[scale=0.43]{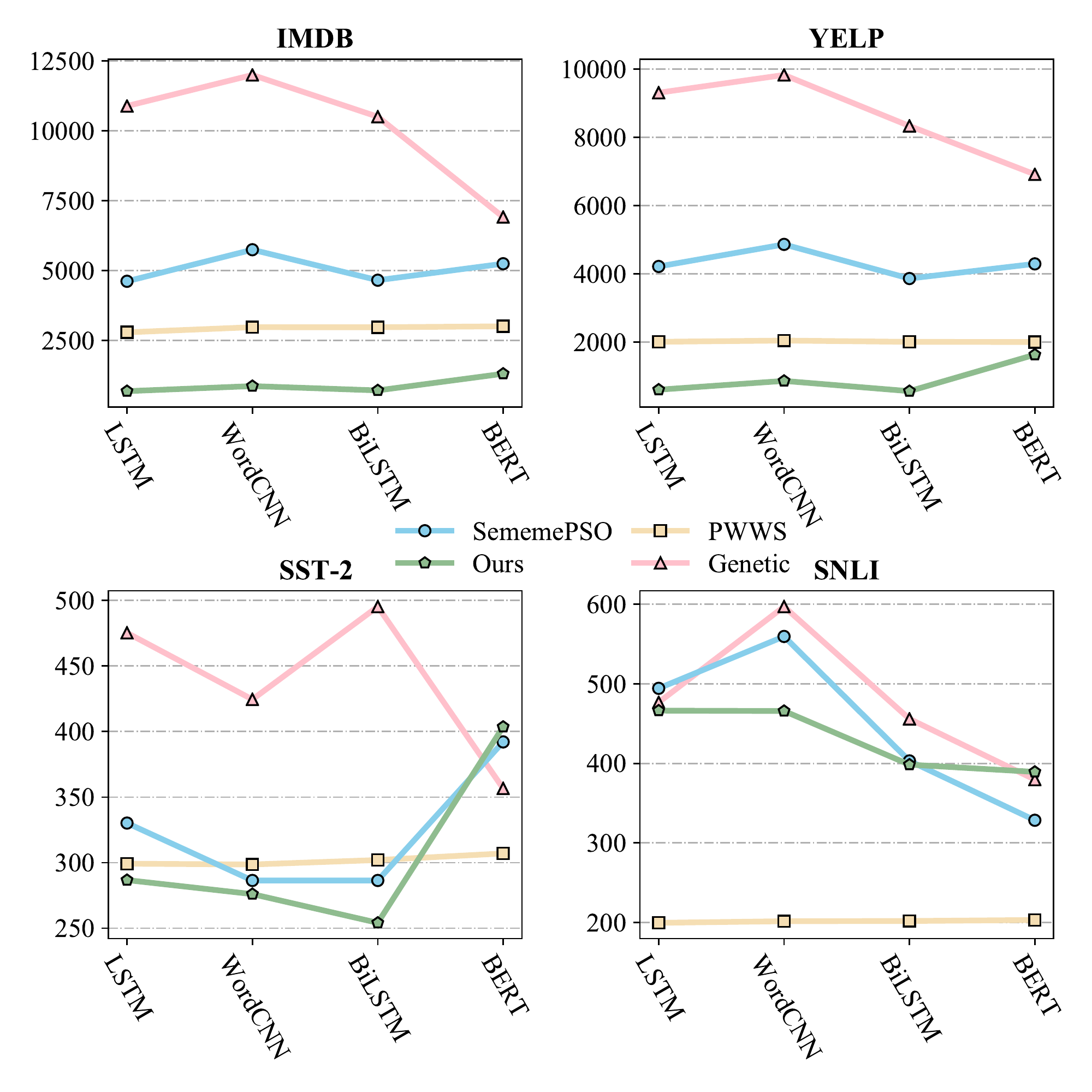}	
	\caption{The number of queries when different attack methods generating an adversarial example. The x-coordinate is the victim model, the y-coordinate is the number of queries.}
	\label{fig:3}
\end{figure}

We use the number of queries for the victim model to reflect the attack efficiency(speed) of the attack methods. Figure \ref{fig:3} visually shows the number of queries(Q) in Table \ref{tabel:3}. Note that our method is significantly efficient than the baseline methods on IMDB and YELP. The number of queries of our method is 1/15-1/3 of the baseline methods, and the attack speed of our method is 3-15 times faster than the baseline methods. Although our attack method is very efficient, our attack success rate has not been affected. On the contrary, it can be seen from Table \ref{tabel:2} that ours attack success rate is higher than the baseline methods in most experiments. As can be seen from Figure \ref{fig:3} and Table \ref{tabel:2} that our method is more efficient than SememePSO and Genetic but less efficient than PWWS on dataset SNLI, which the reason is SNLI is very challenging to attack, the attack success rate of PWWS is much lower than other methods and PWWS  can only successfully attack these simple sentences, which require very few queries. As Table \ref{tabel:2} shows, the attack success rates of the four attack methods vary considerably on dataset SNLI, which explains why our method has a high attack success rate but lower attack efficiency than PWWS.

\subsubsection*{Adversarial Example Quality} 
The quality of the adversarial samples crafted by different attack methods is shown in Table \ref{tabel:3}, note that the evaluation results are calculated by the intersection of the adversarial samples crafted by the four attack methods. We comprehensively evaluate the quality of adversarial samples through modification rate, grammaticality, and language perplexity. It is worth noting that the language perplexity and grammaticality of our attack method are low. This because our method's search strategy does not randomly select synonyms, but according to the degree of similarity calculated by Sentiwordnet between the content word and its synonyms.
But the inadequacy is that the modification rate of our attack method is relatively high.

Finally, Table \ref{tabel:4} shows some adversarial examples generated effectively on BERT by our attack model.

\subsection{Further Analysis} \label{sub:4.7}

\begin{figure}[ht]
	\centering
	\includegraphics[scale=0.44]{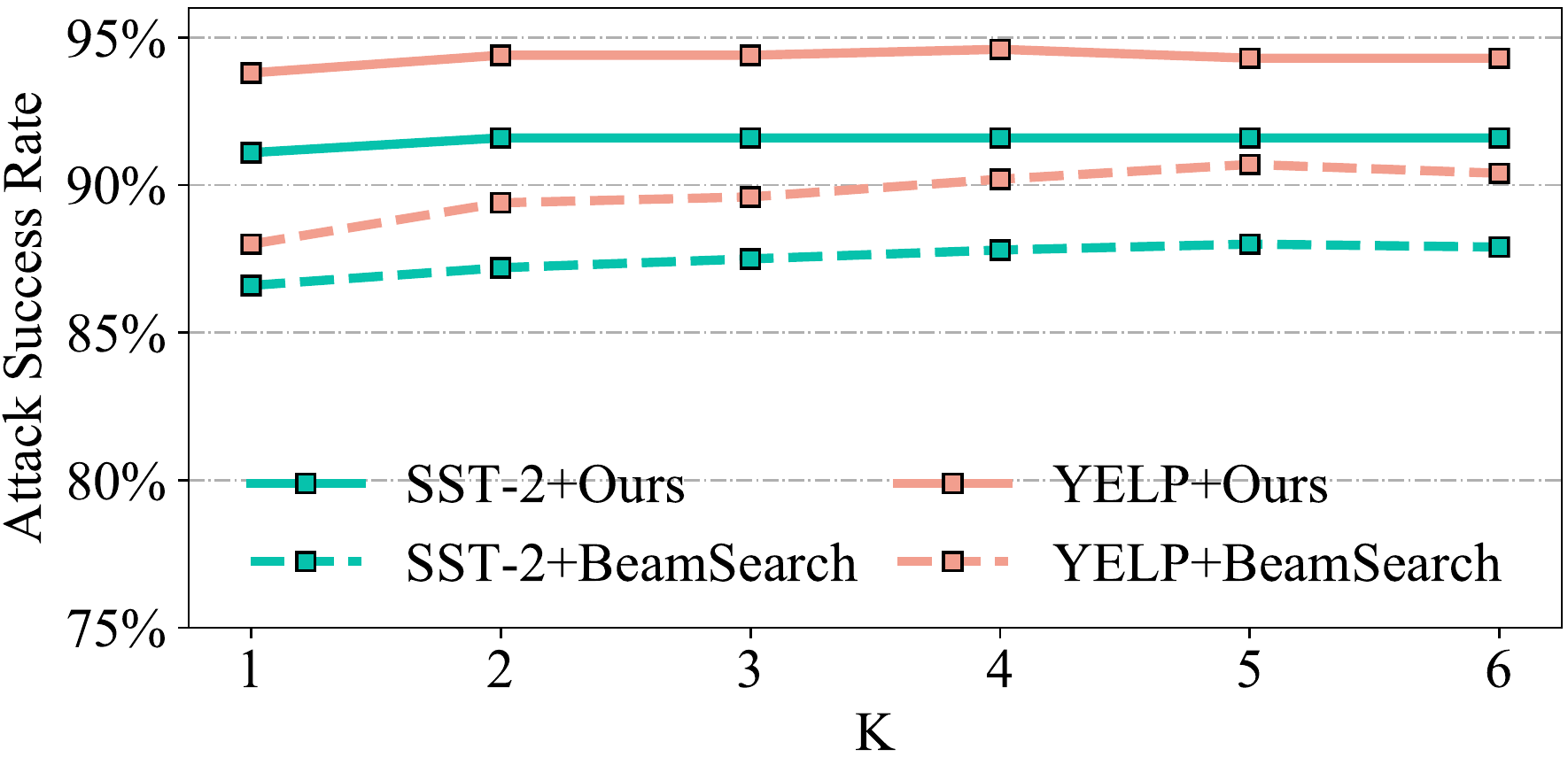}
	\caption{Comparison of attack success rate(\%) between improved beam search algorithm and normal beam search algorithm on SST-2 and YELP. The x-coordinate is beam width, and the y-coordinate is the attack success rate.}
	\label{fig:4}
\end{figure}

We compare the attack success rate of the improved beam search algorithm and the normal beam search algorithm. We choose BERT as the victim model and randomly select 1000 samples which sentence lengths between 10 and 200 from the test set. The comparison results are shown in Figure \ref{fig:4}. 

Figure \ref{fig:4}  visually shows that the improved beam search algorithm outperforms the normal beam search algorithm consistently, and the attack success rate of the improved beam search algorithm has almost converged when the beam width equals 2. It is worth noting that on both datasets, the attack success rate of the improved beam search algorithm with beam width equals one exceeds the attack success rate of the normal beam search algorithm with beam width equals six, which fully proves that our improvement to the normal beam search algorithm against word-level attacks in a black-box scenario is very effective.

\subsection{Transferability}
We use DistilBERT\citep{DBLP:journals/corr/abs-1910-01108} to evaluate the transferability of adversarial examples crafted by SememePSO and our method, and the evaluation metric is the attack success rate. The classification accuracy of DistilBert on IMDB, SST-2, YELP, and SNLI is 90.3\%, 88.3\%, 96.4\%, and 88.3\%, respectively. We only compared with the state-of-the-art attack methond SememePSO because the attack success rate of SememePSO is the closest to our attack success rate. Table \ref{tabel:5} shows that the transferability of the adversarial samples crafted by our attack method is equally matched that crafted by the SememePSO. Therefore, the adversarial samples crafted by our attack method have good transferability.

\section{Conclusion}
In a black-box scenario, the benchmarking attack methods are inefficient, and there is also room for improvement in the attack success rate. In this paper, we propose a novel attack model based on improved beam search algorithm. In addition, we design a different attack strategy and score mechanism. Extensive experiments demonstrate that our attack method is not only more efficient than the baseline methods but has a higher attack success rate. In addition, experiments show that the improvement in efficiency does not cause a decrease in the quality of the adversarial examples, and the adversarial examples have good transferability.

\bibliographystyle{aaai22.bst}
\bibliography{custom2}

%
%

\end{document}